\documentclass[letterpaper, 10 pt, conference]{ieeeconf}  
\usepackage{tikz}
\usetikzlibrary{shapes.geometric, arrows, positioning, shadows, fit}

\IEEEoverridecommandlockouts                              

\title{Brain-Inspired Visual Odometry: Balancing Speed and Interpretability through a System of Systems Approach }
\author{Habib Boloorchi Tabrizi\\
Oklahoma State University\\
Stillwater, OK, USA\\
{\tt\small hboloor@okstate.edu}
\and
Christopher Crick\\
Oklahoma State University\\
Stillwater, OK, USA\\
{\tt\small chriscrick@cs.okstate.edu}
}
\usepackage{graphicx}

\usepackage{subcaption}
\usepackage{amsmath}
\usepackage{amsfonts}

\begin{document}
	
\maketitle
\thispagestyle{empty}
\pagestyle{empty}
	
\begin{abstract}

In this study, we address the critical challenge of balancing speed
and accuracy while maintaining interpretablity in visual odometry (VO)
systems, a pivotal aspect in the field of autonomous navigation and
robotics. Traditional VO systems often face a trade-off between
computational speed and the precision of pose estimation. To tackle
this issue, we introduce an innovative system that synergistically
combines traditional VO methods with a specifically tailored fully
connected network (FCN). Our system is unique in its approach to
handle each degree of freedom independently within the FCN, placing a
strong emphasis on causal inference to enhance interpretability. This
allows for a detailed and accurate assessment of relative pose error
(RPE) across various degrees of freedom, providing a more
comprehensive understanding of parameter variations and movement
dynamics in different environments. Notably, our system demonstrates a
remarkable improvement in processing speed without compromising
accuracy. In certain scenarios, it achieves up to a 5\% reduction in
Root Mean Square Error (RMSE), showcasing its ability to effectively
bridge the gap between speed and accuracy that has long been a
limitation in VO research. This advancement represents a significant
step forward in developing more efficient and reliable VO systems,
with wide-ranging applications in real-time navigation and robotic
systems.
\end{abstract}
	
\section{Introduction}

A "system of systems" is the principle that governs the operation of
the human brain. Each of these subsystems caters to a specific
function. A diverse range of cognitive capabilities is orchestrated by
them when they work together in synchrony. This process is evident in
odometry, which processes spatial navigation and movement. In order
for this skill to emerge, key areas like the hippocampus, entorhinal
cortex, and vestibular nucleus must work together. As a result, we are
able to decipher distance, direction, and motion, which strengthens
our intrinsic navigation skills
\cite{wall2008representation,hitier2014vestibular}.

\begin{figure}
  \centering
  \includegraphics[width=1\columnwidth]{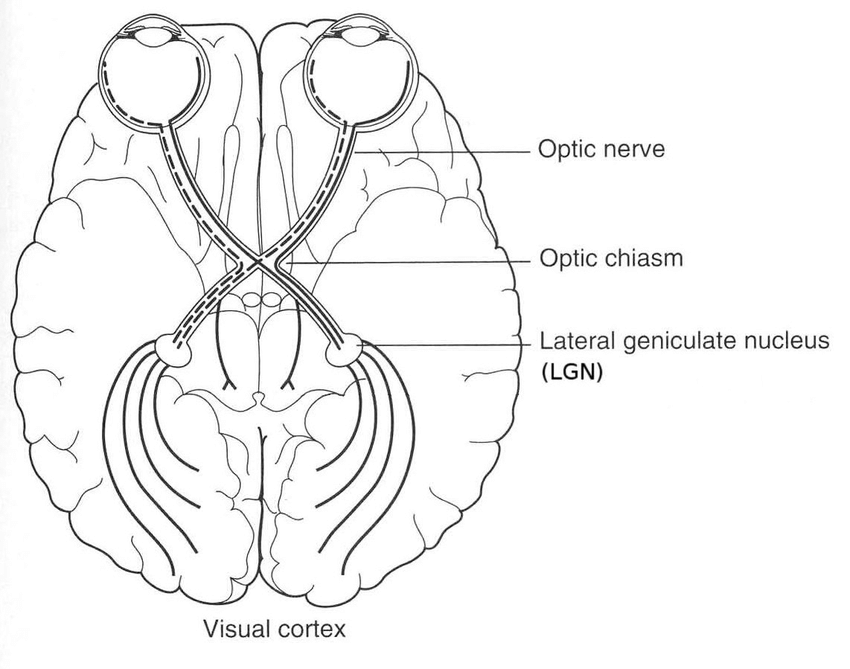}
  \caption{The brain's visual pathway as a ``system of systems''. Each
    component of the visual system specializes in a specific function,
    enabling visual perception
    collaboratively.\cite{eichhorn2007applications}}
  \label{Brain_schematic}
\end{figure}

Figure \ref{Brain_schematic} \cite{eichhorn2007applications}
illustrates the brain's visual system, one of those systems of
systems:

From the retina in the eye, the optic nerve transmits visual
information to the brain.\cite{de2013anatomy}

A data routing element is evident in the optical chiasm. At the optic
chiasm, the optic nerves from each eye cross. In this way, the visual
field from each eye is appropriately divided between the left and
right hemispheres of the brain. \cite{de2013anatomy}

The Lateral Geniculate Nucleus (LGN) is a relay station in the
thalamus that processes visual signals before they reach the visual
cortex. It can be viewed as a preprocessing unit that optimizes and
filters visual data in preparation for higher-level
processing.\cite{de2013anatomy}

It can be compared to the CPU of the visual system, or the main
processing unit. As we perceive our visual environment, the visual
cortex interprets the processed signals from the LGN, located in the
occipital lobe.\cite{jacobson2018visual}

A "system of systems" can be seen in this interconnected flow from the
optic nerve, through the optic chiasm, and LGN, and finally to the
visual cortex. Each component performs a specific function, but works
together to achieve the overarching goal of vision.  The brain
exemplifies the intricate interplay and coordination seen in complex
systems elsewhere.

The brain's system of systems is resilient, and can continue to work
even when one system isn't working properly.  In our model, a fully
connected network receives poses derived from traditional visual
odometry, but provides resiliency so that either system can be changed
without affecting the other. In other approaches, such as
Rovio\cite{rovio15,rovio17}, even a small delay or noisy data from a
sensor or a computation module can compromise the whole system.

Our fully-connected neural network infers from fewer parameters than
convolutional neural network approaches, which often require multiple
frames for inference. The raw pose features obtained from traditional
visual odometry are a sufficient basis for inference in our model.

In this work, we introduce Timestamped Explainable AI. Utilizing
cognitive robotics and innovative AI techniques, this methodology
enhances visual odometry's speed and interpretability by combining
deep learning and traditional visual odometry. Additionally, we can
strike a delicate balance between precision and interpretability at
speeds ranging from 35 to 64 frames per second.

Our brain-inspired "system of systems" is based on three systems.  The
first is the ground truth transformation, which assists in the
training process by making each movement independent of previous
ones. The other two are traditional visual odometry and our deep
learning approach, which are involved in training and testing. Our
results demonstrate the model's interpretability, speed and precision.

\section{Related Work}

Research has been motivated by the complexity and efficiency of the
human brain to develop models that mimic its "system of systems"
structure \cite{jamshidi2017introduction}. Through its modular and
interconnected architecture, the human brain has paved the way for
advances in robotics and computer vision by enabling it to process
vast amounts of information rapidly and accurately. In the vast
expanse of cognitive robotics and artificial intelligence research,
our endeavor stands as a testament to the power of drawing inspiration
from nature's most complex creation: the human brain
\cite{berco2019recent}. 

When viewed through the lens of the brain's "system of systems"
structure, we have developed a visual odometry approach that is both
efficient and interpretable.  The brain's sophisticated network of
interconnected ``system of systems'' \cite{ruehl2022human} provides a
perspective that helps unravel the multi-faceted complexities and
interconnectedness foundational to our cognitive processes
\cite{axenie2015cortically}.

Venturing deeper into visual processing, the retina's indispensable
role shines through. The ganglion cells' axons converge to form the
optic nerve, predominantly connecting to the thalamus, especially the
lateral geniculate nucleus (LGN), and the superior colliculus. The LGN
stands as a pivotal relay, ensuring the retina's visual information
seamlessly reaches the cortex \cite{hitier2014vestibular}.


Interpretability and explainability are fundamental to deep learning
\cite{gilpin2018explaining}. The former looks at the consequences of
tweaking parameters, while the latter traces the causes of AI
decisions. The integration of these ideals with visual odometry
presents some challenges From navigating the intricate web of modern
VO systems to understanding the limitations of traditional
methodologies, there are both opportunities and challenges.

By using visual odometry, a camera's position can be estimated. Using
subsequent frames, the camera's orientation and position are
calculated. After collecting a sequence of images from the camera,
motion is calculated based on a reference frame to estimate the
camera's position. A feature is identified and matched between
successive images to accomplish this. Robots can use odometry to
determine their position and orientation without depending on external
sensors, making it an important tool in robotics and computer
vision\cite{cadena2016past,huang2017visual,valencia2018mapping}.
	
We are developing a system for monocular visual odometry in this
research. Several factors influenced our decision. It is true that
integrating IMUs or GPS can provide additional information, but it
also introduces additional error sources and complexities in system
design and calibration
\cite{rovio15,rovio17,gui2015mems,maimone2007two}.  It is our goal to
create a method for monocular visual odometry that is both
computationally efficient and easy to interpret.

A feature-based method, ORB-SLAM, keeps track of the camera pose by
detecting features in the images. Detecting features can be
challenging in areas with low-contrast textures and dynamic
objects. ORB-SLAM overcomes this problem by using a variety of
techniques, including feature matching, key frame selection, and loop
detection. As a result of these methods, the system is able to
accurately determine the camera's pose in challenging
environments. Furthermore, the system is capable of detecting loop
closure, which allows it to accurately determine the trajectory of the
camera. \cite{orb-slam1,orb-slam2}
	
The disadvantage of ORB-SLAM is that it can be computationally
expensive, since it requires a large number of feature points to be
detected to accurately calculate the camera's pose
\cite{mur2015orb}. Additionally, since the system relies on feature
detection, errors can occur if the features are not correctly
identified. Inaccurate results can compromise the accuracy of the
visual odometry \cite{yang2020multi}.

DSO (Direct Sparse Odometry) is a direct method for estimating the
6-DoF camera pose\cite{engel2018direct}. Instead of detecting
features, it focuses on the raw intensity of the images, as opposed to
traditional feature-based methods. This method involves taking two
images of a scene and calculating the camera's relative 3D motion. In
mobile applications and robotics, DSO is a real-time algorithm that
runs on a single processor. Due to its robustness to motion blur, this
algorithm does not require prior knowledge of the scene. The main
disadvantage of DSO is that it is sensitive to lighting conditions, so
it is not suitable for low-light environments. In addition, it is less
reliable when dealing with fast motion and high-frequency textures.

As a lightweight method, MonoVO is computationally efficient. Because
of this, it is fast, but it can also result in lower accuracy than
methods that require more computational resources. Because MonoVO uses
an inverse depth parameterization for camera motion, it is able to
achieve this efficiency more easily than other representations. The
limited resolution of the parameterization, however, can also reduce
the accuracy of the estimates\cite{delmerico2018benchmark}.

Visual odometry has benefited greatly from deep learning methods in
recent years \cite{wang2017deepvo}. The underlying features of the
task have been learned using deep learning methods to achieve better
results than traditional approaches. Deep learning methods can improve
visual odometry accuracy, speed, and
generalization\cite{muller2017flowdometry}. A deep learning model can
also be used to gain a deeper understanding of a scene's underlying
dynamics in addition to predicting future motion. As a result of this
prediction, the hardware will be burdened with additional complexity
\cite{saputra2018visual,he2020review}.
	
A recent trend in visual odometry is the integration of deep learning
techniques \cite{pandey2021leveraging}. Sequence handling can be
accomplished with LSTMs without having to store entire image series
\cite{zhu2022deepavo}. In addition, these approaches require a high
level of training and suffer from cumulative errors As a result, they
are not suitable for practical, large-scale
applications\cite{xue2019beyond, chen2023learning}.

Our novel approach combines deep learning and traditional visual
odometry techniques as separate systems and focuses on speed and
accuracy. Using visual odometry in our deep neural network enhances
interpretability by breaking down the model into individual degrees of
freedom.
	
Since last moves have no effect on estimating pose in current moves,
our model is robust even in scenarios with fast or abrupt
movements. Our goal is to surpass the limitations of current visual
odometry models by eschewing the LSTM approach and focusing on a fast
and interpretable model, thus advancing robotics and autonomous
navigation.
	
\section{Technical Description}

We aim to develop a platform that is easy to understand and fast by
taking inspiration from the way the brain's distinct regions work
together harmoniously for faster performance. The system is divided
into three subsystems: ground truth data transformation, visual
odometry, and deep learning. By increasing visual odometry's
modularity and deep learning's robustness, we are able to mimic the
brain's complex "system of systems."

\begin{figure*}[t]
  \centering
  \includegraphics[width=2\columnwidth]{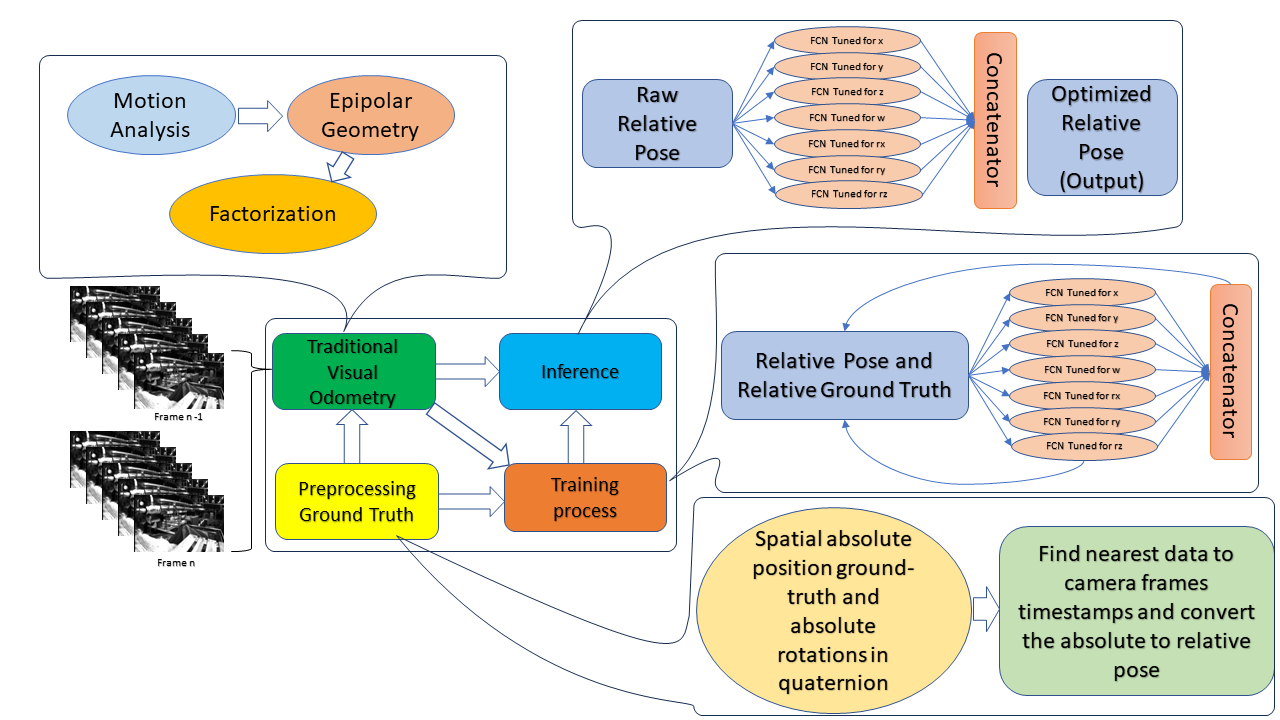} %
  \caption{Comprehensive pipelined traditional visual odometry as a
    "system of systems".  Geometry is determined using the essential
    matrix, while the motion is calculated using optical flow based on
    the Harris \cite{harris1988combined} method. Our ground truth
    system has been modified to convert the absolute pose to a
    relative one. By using the ground truth as a target, the
    optimization process can calculate loss with greater accuracy. We
    have incorporated several specialized fully connected networks
    during the training phase, each fine-tuned for a particular degree
    of freedom. Though the structures of inference and training appear
    similar, it is important to note that the inference stage does not
    include backpropagation and its gradient setting is
    disabled. After each degree of freedom FCN has been deactivated, a
    segment of the training procedure focuses on the concatenation of
    the individual fully connected networks for each degree.}
  \label{methodflow}
\end{figure*}

Figure \ref{methodflow} depicts the system of systems, capturing not
only the essence of traditional visual odometry, but also the
flexibility of our deep neural network. We intentionally left our
platform's design customizable, emphasizing the importance of
adaptability in response to varying inputs, similar to the
adaptability of the brain. Yet, comprehensive details of the fully
connected networks used in our experiments are available.

To interpret raw movements, our traditional visual odometry subsystem
employs sparse optical flow, which is analogous to the brain's spatial
processing. Our neural network subsystem relies on these movements,
harmonized with the adjusted ground truth for each frame pair.

We have crafted a machine learning platform that emphasizes
modularity, interpretability, and adaptability by embracing the
brain's "System of Systems" architecture.
 
With sparse optical flow, we capture raw movements using traditional
visual odometry. Our neural network uses these movements as variables,
along with the adjusted ground truth generated for each pair of
frames.
	
\subsection{Ground-truth transformation}

We didn't use absolute frames since we didn't want to be dependent on
the last move. As a result, we would not have to use an LSTM or
another reinforcement technique and the hardware would not be burdened
as much and the complexity would be reduced.
 
When evaluating movements, we emphasize relative frames over absolute
frames. A global coordinate system defines an absolute frame as the
camera's position and orientation. Alternatively, a relative frame
refers to the camera's position and orientation as compared to its
previous position.
	
\begin{figure}
  \centering
  \includegraphics[width=1\columnwidth]{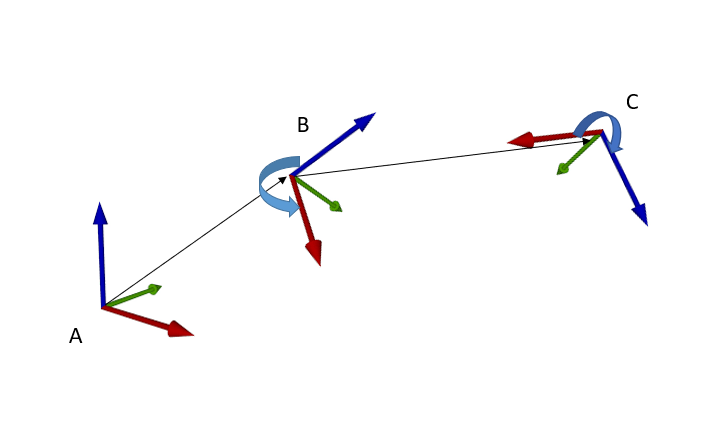}
  \caption{Transitioning from frame A to frame B, and then to frame C,
    the rotation results in a completely different space where all
    axes, areas, and origins can change. For instance, axis X in frame
    A could potentially become Axis Z in frame C.}
  \label{fig1}
\end{figure}

When rotated or translated, each movement occupies a new space (see
Figure \ref{fig1}).
	
It is necessary to factor in a transformation for each movement, which
encompasses both rotation and translation matrices
\cite{frank2017modern,murray2017mathematical,cao2019novel}.
	
We consider rotation and translation matrices whenever we perform a
movement:
	
$$
T = \begin{pmatrix}
  R & t\\
  o & 1
\end{pmatrix} = \begin{pmatrix}
  r_{11} & r_{12} & r_{13} & p_1\\
  r_{21} & r_{22} & r_{23} & p_2\\
  r_{31} & r_{32} & r_{33} & p_3\\
  0 & 0 & 0 & 1
\end{pmatrix}
$$

In this matrix, $R$ signifies a rotation in the Special Orthogonal Group
(SO(3)), while t is a column vector in $\mathbb{R}^3$. The motion of a
rigid body can be depicted by a set of all 4x4 real matrices, known as
the Special Euclidean group (SE(3)). The element $T$ in SE(3) can be
portrayed as the pair $(R,t)$.
	
In order to determine the elements of the inverse transformation
matrix, we multiply a matrix with its inverse, as demonstrated below:
$$
T^{-1} = \begin{pmatrix}
  R & p\\
  O & 1\\
\end{pmatrix}^{-1} = \begin{pmatrix}
  R^T & -R^T p\\
  O & 1\\
\end{pmatrix} \in SE(3)
$$
We can see that:
$$
T_{bc} = T_{bs}T_{sc} = {T_{sb}}^{-1}T_{sc}
$$

Bearing these considerations in mind, we've been able to transform the
ground truth from absolute rotation and translation to relative
rotation and translation.
	
\subsection{Traditional Visual Odometry}

As we endeavor to replicate the brain's "system of systems"
architecture, we recognize that each subsystem must function
efficiently in order to ensure that the overall system runs
efficiently. Similarly, our Traditional Visual Odometry subsystem is
optimized for maximum speed and efficiency, just as the brain is
capable of handling vast amounts of data quickly.

We use sparse optical flow to derive the essential matrix linking
corresponding points between two images. In low-texture environments,
this method can track a select set of points over time. With fewer
points being monitored, the processing speed is amplified and no
calibration or high-end cameras are required to process
images. Essentially, this ensures that our Visual Odometry will not
become a bottleneck, providing the necessary pace to the overarching
system.

In order to detect corners precisely, we utilize the Harris technique
\cite{harris1988combined}, while the Shi-Tomasi \cite{shi1993good}
down-sampling approach for patches ensures robustness against noisy
images.  Singular value decomposition is used to extract rotation and
translation matrices for our pose recovery strategy.

In our design, Traditional Visual Odometry is principally used to
provide insight into the transformation between two consecutive frames
instead of a key frames. Because there are more frames in each second,
even if we have an extreme error in two frames, the mean of the
relative pose error will be reduced.

\subsection{Deep Learning Approach}
	
We have developed an innovative deep learning approach that aligns
with the brain's modular functionality by drawing inspiration from its
"system of systems" architecture. Several distinct regions of the
brain are dedicated to specialized tasks, illustrating the importance
of optimizing subsystems for precise functionality. We also fine-tune
every degree of freedom individually to ensure that each component is
optimized to its maximum potential.

As a result of the inherent differences in movement patterns, such as
stark rotations around the $x$ and $y$ axes compared to the more
stable rotations around the $z$ axis, we have opted for a Fully
Connected Network (FCN) over the more commonly adopted Convolutional
Neural Network (CNN). Since the FCN does not process entire images, it
promises a faster processing speed due to its fewer parameters. We
made this choice in order to ensure that our approach remains swift
and efficient, reflecting how individual brain subsystems contribute
to the overall efficiency and speed of cognitive processes.

As each degree of freedom is trained independently, we are able to
perform a granular analysis, so that we can determine if each
pre-trained neural network branch is compatible with movement
patterns, and determine whether a different architectural approach is
required.

When compared with end-to-end analyses, this modular approach not only
enables a custom-tailored model for each movement and degree of
freedom, but also offers a holistic perspective. We are able to gain a
deeper understanding of both absolute trajectory and relative pose by
enhancing the interpretability of our model.

The modularity and customization of our activation functions,
including Relu, Leaky-Relu, Elu, Tanh, and Sigmoid, demonstrates our
commitment to modularity and customization. Based on the specific
requirements, we can target either global error reduction (absolute
trajectory errors - ATE) or local error reduction (relative pose
errors - RPE), as shown in table \ref{RPE}.
	

To describe this mathematically, let's denote each degree of freedom
as $d_i$, where $i=1,2,\ldots,n$ and $n$ is the total number of
degrees of freedom. The model for each degree of freedom can be
represented as $M_i(d_i)$. Rather than learning a combined model
$M(d_1,d_2,\ldots,d_n)$ jointly, we optimize each $M_i$ individually
and then combine them into an integrated model. This can be formulated
as:

\begin{equation}
  M(d_1, d_2, \ldots, d_n) = \bigcup_{i=1}^{n} M_i(d_i)
\end{equation}
	
By optimizing each $M_i(d_i)$ separately, we can examine the
contribution of each degree of freedom to the overall motion in a more
detailed way, thus obtaining a more accurate model. This process
results in greater transparency and comprehension of the underlying
movements and poses. The comparison with the end-to-end approach, as
shown in table \ref{RPE}, validates that our method provides a more
precise understanding of the influence of each degree of freedom.  A
model based on individual optimization can be used to examine in depth
how each degree of freedom contributes to overall motion, resulting in
a more accurate and transparent model. Comparing our method with the
end-to-end approach, as presented in table \ref{RPE}, further
reinforces the precision and understanding we provide regarding the
influence of each degree of freedom.
	
At the end we used pretrained branches and as we can see in Figure
\ref{methodflow}, we used inception to get better result for the
traine model. In other words, we used different modules based on our
customized desired result for less Relative Pose Error.
	
\section{Experimental Results}
	
In our research, we have merged traditional visual odometry techniques
with an analytical method for pinpointing and tracking points of
interest through patch similarity. This synergy was anticipated to
yield a fast and efficient method, particularly valuable for
low-performance embedded devices, a notable contribution of our study.

Remarkably, each processing step in our system requires only 15.625 to
31.25 microseconds in our modest test environment. When applied in
real-world scenarios on a standard Core i7 laptop, this efficiency
translates into an exceptional frame rate, enabling real-time or even
accelerated processing, as we demonstrate.

Our method's pipelined structure also led us to test the essential
matrix in addition to the fundamental matrix within the epipolar
geometry framework. The results averaged out to show no significant
disparity in speed or precision between the two. However, the
essential matrix mode, utilizing specifications from the right camera
of the EuRoC dataset \cite{euroc}, exhibited about 0.1 meters less
error compared to the traditional odometry mode. This error margin
diminished even further when augmented with our fully connected neural
network.

\begin{figure}
  \centering
  \includegraphics[width=1\columnwidth]{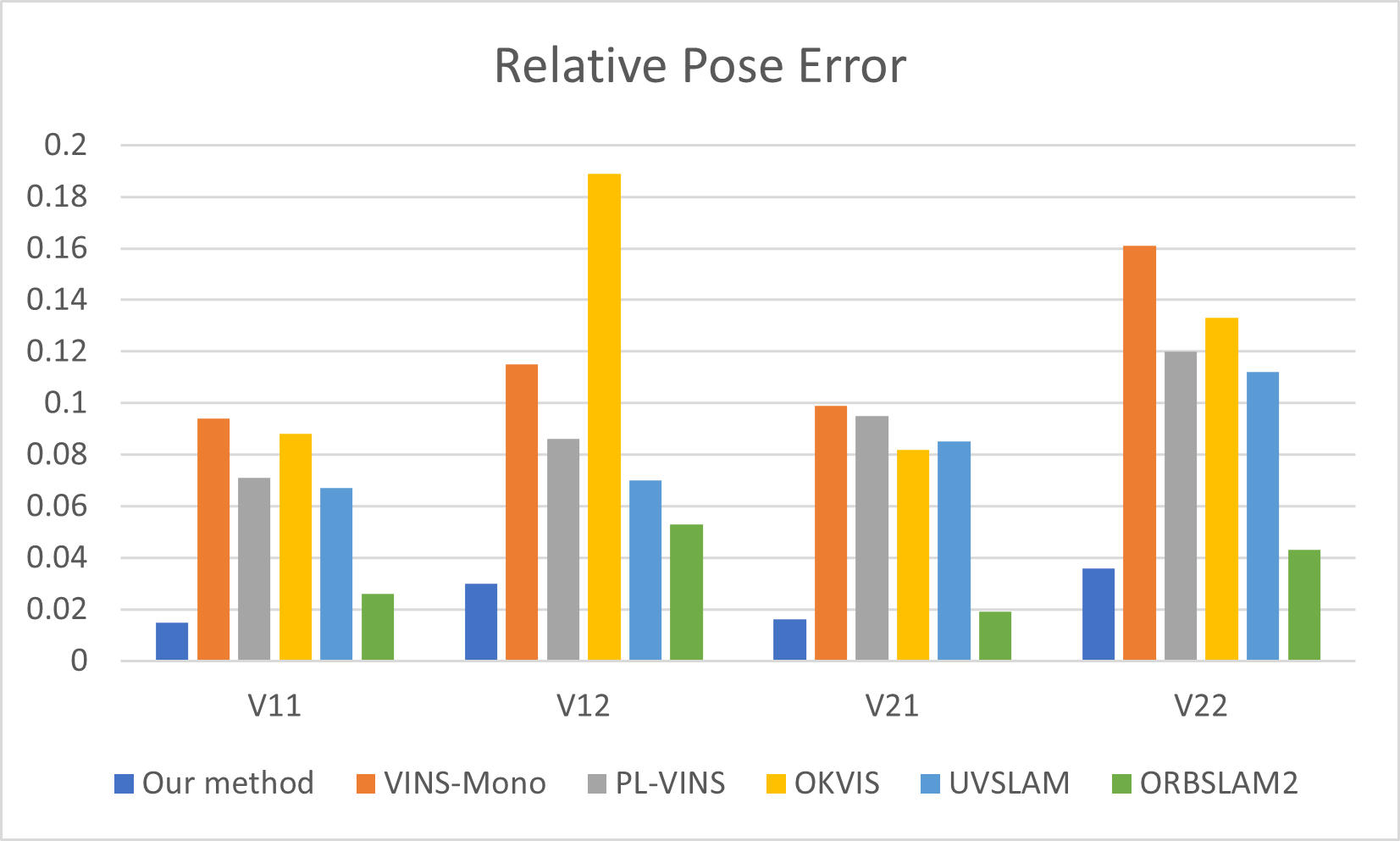}
  \caption{ We compared our method with other methods
    \cite{chen2019stereo,lim2022uv} in different scenarios of the
    Vicon room of the Euroc \cite{euroc} dataset for relative pose
    error. Throughout each scenario, the blue color represents our
    method's error.  }
  \label{RPE_Comaprison}
\end{figure}

Further experiments conducted with the EuRoC MAV dataset \cite{euroc}
underscored our approach's superiority. As depicted in Figure
\ref{RPE_Comaprison}, our method consistently outperformed various
contemporary counterparts, including those integrating inertia as an
observer \cite{chen2019stereo,lim2022uv}. These results affirm our
method's ability to maintain or surpass accuracy while significantly
enhancing computation speed.

Our approach's decomposition into smaller subproblems facilitated more
manageable training and optimization phases. This modular design
allowed for the independent training of different components, such as
the feature extractor and the pose estimator. Such decomposition
enabled training on extensive datasets, which would have been
impractical with a singular, monolithic model.
	
Using this refined method, we achieved state-of-the-art results on the
EuRoC dataset \cite{euroc}.  Particularly notable was an Absolute
Trajectory Error (ATE) of 0.0149 on the V11 sequence, as shown in
Table \ref{ATE}. The efficacy of our modular architecture and
decomposition strategy was evident in these results, detailed in Table
\ref{RPE}.
	
\begin{table*}[ht]
  \centering
  \caption{RPE Translation for different activation functions}
  \small
  \setlength{\tabcolsep}{4pt}
  \begin{tabular}{|c||c|c|c|c|c|c|c|}
    \hline
    Activation & RPE Trans. X & RPE Trans. Y & RPE Trans. Z & RPE Trans. & RPE Rot. RX & RPE Rot. RY & RPE Rot. RZ \\
    \hline
    Leaky ReLU & 2.0151 & 1.0233 & 2.2991 & 1.& 0.6825 & 0.5212 & 0.6468  \\
    \hline
    SELU & 2.2198 & 1.3453 & 1.7351 & 1.8026 & 0.4322 & 0.4109 & 0.7000 \\
    \hline
    Tanh & 1.2642 & 1.4221 & 1.4629 & 1.3857 & 0.6186 & 0.7383 & 0.6360 \\
    \hline
    ReLU & 3.4905 & 2.8742 & 3.2816 & 3.2256& 0.4187 & 0.3926 & 0.6707  \\
    \hline
    Sigmoid & 1.6106 & 1.5485 & 2.0732 & 1.7598 & 0.6321 & 0.3583 & 0.6258\\
    \hline
  \end{tabular}
  \label{RPE}
\end{table*}
	
\begin{table*}[ht]
  \centering
  
  \caption{ATE for different activation functions}
  \small
  \setlength{\tabcolsep}{5pt}
  \begin{tabular}{|c||c|c|c|c|}
    \hline
    Activation & ATE Trans. X & ATE Trans. Y & ATE Trans. Z & Mean ATE \\
    \hline
    Leaky ReLU & 2.0151 & 1.0233 & 2.2991 & 1.7792 \\
    \hline
    SELU & 2.2198 & 1.3453 & 1.7351 & 1.7667 \\
    \hline
    Tanh & 1.2642 & 1.4221 & 1.4629 & 1.3831 \\
    \hline
    ReLU & 3.4905 & 2.8742 & 3.2816 & 3.2154 \\
    \hline
    Sigmoid & 1.6106 & 1.5485 & 2.0732 & 1.7441 \\
    \hline
  \end{tabular}
  \label{ATE}
\end{table*}
	
The optimization of activation functions based on specific task
requirements was another aspect of our research. This exploration
demonstrated that different activation functions can influence model
performance variably, contributing to both the model's speed and
interpretability.

In conclusion, our findings illustrate that visual odometry (VO)
models can achieve both interpretability and speed through a
well-crafted, brain-inspired hybrid approach. Our system, with
processing speeds ranging between 15.625 to 31.25 microseconds and
high frame rates on standard computing hardware, sets a new benchmark
in the field. It heralds a future where speed, accuracy, and
transparency in VO models coexist harmoniously.

	\section{Conclusion}
	
Our endeavor stands as a testament to the power of drawing inspiration
from nature's most complex creation: the human brain, in the vast
expanse of cognitive robotics and artificial intelligence research. We
have developed a visual odometry approach that is both efficient and
interpretable when viewed through the lens of the brain's "System of
Systems" structure.

We combine traditional visual odometry techniques with the
capabilities of a fully connected network, embodying the modularity
and interconnectedness of neural systems. Through this fusion, we have
developed a methodology that is not only explainable but also
demonstrates remarkable performance metrics. We believe that our
approach will be valuable in real-world applications, particularly
because of the speed enhancements achieved by our system, clocking
impressive frame rates on standard hardware.

In addition, our modular method ensures a level of granularity and
clarity in the results that is often lacking in contemporary research
due to the fact that each degree of freedom is treated
individually. In our experiments, we have consistently outperformed
several benchmarks, further demonstrating the efficacy of our
approach.

In essence, we have paved a way for future research in this area by
combining cognitive robotics with cutting-edge artificial intelligence
techniques. As a result of our work, we believe that future endeavors
will be guided towards a harmonious blend of speed, accuracy, and
interpretability. As we move further into the era of artificial
intelligence, our research emphasizes the profound potential in
looking back to nature for inspiration, ensuring a future in which
machines function with a precision and transparency reminiscent of
human thinking.

\bibliographystyle{ieeetr}
	
\bibliography{example}  
	
\end{document}